\documentclass[]{article}
\usepackage[affil-it]{authblk}
\usepackage[a4paper, total={6in, 8in}]{geometry}
\usepackage{graphicx}
\usepackage{pgfplots}
\usepackage{booktabs} 
\usepackage{graphicx}
\usepackage{pgfplots}
\usepackage[all]{nowidow}
\usepackage[utf8]{inputenc}
\usepackage{tikz}
\usepackage{fancyhdr}
\usepackage{dblfloatfix}
\usepackage{float}
\usepackage{textgreek}
\usepackage{multicol}
\usepackage{algorithm}
\usepackage{relsize}
\usepackage{mathtools}

\usepackage{hyperref}
\usepackage{subfig}
\usepackage{verbatim}
\usepackage{algorithmic}
\usepackage[inline]{enumitem} 
\usepackage{setspace}

\title{\textbf{Variational EP with Probabilistic Backpropagation for Bayesian Neural Networks}}
\author{Kehinde Olobatuyi%
	\thanks{Electronic address: \texttt{k.olobatuyi@campus.unimib.it}; Corresponding author}}
\affil{Statistics and Mathematical Finance,\\ University of Milano-Bicocca, Italy}

\date{}

\definecolor{blue}{HTML}{1F77B4}
\definecolor{orange}{HTML}{FF7F0E}
\definecolor{green}{HTML}{2CA02C}

\pgfplotsset{compat=1.14}

\newcommand\MEETtitle[1]{\Large \bf \hskip2.25pc \parbox{.8\textwidth}{ \noindent%
		\large \bf \begin{center} #1 \end{center}\rm } \vskip.1in \rm\normalsize }

\newcommand\MEETauthor[1]{\hskip2.25pc \parbox{.8\textwidth}{ \noindent%
		\normalsize \bf \begin{center} #1 \end{center}\rm } \vskip-1pc }

\let\title\MEETtitle
\let\author\MEETauthor
\let\affil\MEETaddress

\begin{document}

\maketitle

\begin{abstract}
I propose a novel approach for nonlinear Logistic regression using a two-layer neural network (NN) model structure with hierarchical priors on the network weights. I present a hybrid of expectation propagation called Variational Expectation Propagation approach (VEP) for approximate integration over the posterior distribution of the weights, the hierarchical scale parameters of the priors and zeta. Using a factorized posterior approximation I derive a computationally efficient algorithm, whose complexity scales similarly to an ensemble of independent sparse logistic models. The approach can be extended beyond standard activation functions and NN model structures to form flexible nonlinear binary predictors from multiple sparse linear models. I consider a hierarchical Bayesian model with logistic regression likelihood and a Gaussian prior distribution over the parameters called weights and hyperparameters. I work in the perspective of E step and M step for computing the approximating posterior and updating the parameters using the computed posterior respectively.
\end{abstract}

\section{Introduction}
Bayesian methods have several benefits and have been used to solve many complex models. In particular, their uniform treatment of uncertainty at all cost makes them a unique methods. One of the breakthrough of Bayesian methods is the application to Neural Networks. Neural Networks (NNs) has seen a resurgence in the area of supervised learning problems. In their typical usage, neural networks are highly expressive models that have the capability of learning complex functions approximations from input/output examples (\cite{hornik1989}). Some of the achievements is due to the ability to train them on massive data sets with stochastic optimization (\cite{bottou2010}), and the updates of the parameters called weights using the backpropagation (\cite{rumelhart1986}). Coupled with faster machines, larger datasets, and innovation such as regularization techniques (dropout) (\cite{srivastava2014}) and rectified linear unit for transformation to nonlinear functions (\cite{nair2010}), have contributed to the successes of NNs on tasks such as speech recognition (\cite{hintonet2012}; \cite{Hannunetal2014}), natural language processing (\cite{collobert2008}; \cite{sutskever2014}) and computer vision (\cite{krizhevsky2012}). However, there are still some challenges in learning NNs due to the following reasons; first, working with large data sets will result to specific tuning of many hyperparameters, such as learning rate, momentum, and weight decay which are all layer-dependent. Second, NNs only output point estimates of the weights in the network. Networks with point estimates do not account for uncertainty in the parameters. As a result, in many cases these weights may perform poorly. It is desirable to produce uncertainty estimates along with the predictions. Moreover, there often exists a problem of overfitting and underfitting (too complex and too simple). A Bayesian approach to neural network has been used in the literature to avoid these pitfalls of the stochastic optimization (\cite{mackay1992c}). Bayesian techniques, in practice can infer values for hyperparameters by marginalizing out the parameters automatically out of the posterior distribution. Furthermore, Bayesian techniques provide for uncertainty of estimates into prediction. Bayesian approaches are often robust to overfitting by averaging over parameter values of a single point estimate. Many approaches have been proposed for Bayesian learning of neural networks such as Laplace approximation (\cite{mackay1992c}), Hamilton Monte Carlo (\cite{neal1995}), Expectation Propagation (\cite{Jylankietal2014}), and Variational inference (\cite{hintonandcamp1993}). However in particular, EP has not seen widespread adoption due to the lack of simplicity in computation of normalizing factor by integrating out the parameters which results in rigorous analytical computation when the likelihood and the prior distribution are not in the same quadratic exponential family. A notable prevalence exception is the scalable variational inference approach by \cite{grave2011vi}. However, this method performs poorly in practice due to noise from Monte Carlo approximation within the stochastic gradient computation. Moreover, the scalable solution based on EP focuses on networks with binary weights, while its extension to continuous weights becomes unsatisfying it does not capture the uncertainty in posterior variance. Gaussian priors may seem to be an inappropriate choice for the input layer weight of a feed-forward neural network (NN) because allowing, \textit{a priori}, a large weight $w_j$ for a potentially irrelevant variable $x_j$ may deteriorate the predictive power of the model. This behavior is analogous to a linear model as the input layer weights associated with each hidden unit of the multilayer perceptron (MLP) network is seen as a separate linear models whose outputs are transformed nonlinearly in the next layer. Integrating the posterior uncertainty over the unknown input weights alleviates the potentially harmful effects of the irrelevant variables. However, it may not be sufficient if the total number of input variables or features grows proportionately large with the number of observations. An alternative strategy to suppress the effect of the irrelevant features is to combine linear model with sparsity-promoting priors using general activation functions and interaction term between the hidden units (\cite{Jylankietal2014}). A similar approach has been to apply hierarchical automatic relevance determination (ARD) priors (\cite{mackay1992c}; \cite{neal1996}), where individual Gaussian priors are assigned for each weights $w_j = \mathcal{N}(0,\tau_{l_j})$, with separate variance hyperparameters $\tau_{l_j}$ regularizing the group of weight associated to each feature. \cite{mackay1995a} described an ARD approach for NNs where the integration over the point estimate of the relevance parameters $\tau_{j}$ and other model hyperparameters such as the noise parameters are approximated using the Laplace's method to compute the marginal likelihood estimate. Moreover, \cite{neal1996} approached the same problem from the angle of stochastic approximation such as  Markov chain Monte Carlo (MCMC) where the approximate integration is performed over the posterior uncertainty of all the model parameters including $w_j$ and $\tau_{l_j}$. Various computationally efficient algorithms have been proposed for determining marginal likelihood based point estimates for the relevance parameters (\cite{tipping2001}; \cite{qietal2004}; \cite{wipfandnagarajan2008}). However, these are in connection with linear models and the point-estimate based methods may suffer from overfitting because the maximum a posteriori (MAP) estimate of $\tau_{l_j}$ may be close to zero also for relevant features as demonstrated by \cite{qietal2004}. This can also be applied to the implementation of infinite neural networks using Gaussian process (GP) priors when separate hyperparameters controlling the nonlinearity of each input are optimized (Williams, $1998$; \cite{rasmussenandwilliams2006}). A surrogate for ARD priors based on sparsity-favoring priors such as Laplace prior (\cite{seeger2008pro}) and spike and slab prior (\cite{joseetal2008}, \cite{jose2010}) have been widely presented for linear models. Expectation propagation has been used for this methods to efficiently integrate over the analytically intractable posterior distributions. The advantage of these methods over ARD prior is that they do not suffer from overfitting. This is due to the approximate integration over the posterior uncertainty resulting from a sparse prior on the weights. Expectation propagation provides a useful alternative to MCMC for approximating the integration as a result of its computational efficiency and accuracy in many practical applications (Nickish and Rasmussen, $2008$; \cite{jose2010}). In nonlinear regression, sparsity favoring Laplace priors have been considered for NNs (Williams, $1995$), where the inference is performed by Laplace approximation. However, Laplace approximation suffers from the problem of discontinuity i.e. the curvature of the log-posterior density at the posterior mode may not be well defined for all types of prior distributions, including the Laplace distribution whose derivatives are not continuous at the origin (Williams, $1995$; \cite{seeger2008pro}). A successful implementation of this algorithm requires some additional approximations (Williams, $1995$). In contrast, EP provides a straightforward implementation due to the propagation of moments of the priors terms with respect to a Gaussian measure. Another properties of Laplace approximation is that it approximates the posterior means of the unknown quantities with the maximum a posteriori (MAP) estimates and their posterior covariance with the negative Hessian of the posterior distribution at the mode. This setup may underestimate the overall uncertainty of the unknown variables which may lead to poor predictive performance of the skewed posterior distributions (Nickish and Rasmussen, $2008$) or multimodal (Jylanki et al., $2011$). Furthermore, in typical practical NN applications, the MAP may differ significantly from the posterior mean when there are many unknowns compared to the number of observations. 
\subsection{Main Contribution}
The goal of this work is to study the Expectation Propagation and Variational Inference methods from an intertwined viewpoints and a different divergence standpoint. We make a new unification of EP and VB algorithms which makes EP less analytically rigorous for the hierarchical Bayesian framework. Variational EP (VEP) incorporates the propagation algorithm of EP at the VB updating stage. VEP focuses on the core limitation of expectation propagation algorithm while generalizing the Variational inference algorithm. The most vital part of the EP and VB algorithm is the minimization of the Kullback-Leibler divergence. This new approach to both EP and VB method is called the Variational Expectation Propagation (VEP) algorithm. VEP has several advantages such as the generalization of the Variational Inference algorithm by incorporating the refining strategy of EP into VB. Moreover, the refinement of the prior through the data instead of fixed contribution of the prior to the approximate posterior in VB would be expected to improve the accuracy of VB although at the expense of the global update rule. Additionally, VEP connect a path or mediates between EP and VB algorithms. VEP breaches the rigorous analytical problems of EP by transiting from the VI algorithm with an augmentation. Additional, solving the intractable tilted posterior distribution of EP with a VB approach. This leads to working in the perspective of hybridizing the EP and VB to approximate hierarchical Bayesian models. We provide some theoretical framework that establish the connecting linkage between EP and VB. Furthermore, we investigate the sparse linear models into nonlinear regression following the strategy by \cite{Jylankietal2014} which combines the sparsity favoring priors with a two-layer regression models. This aims to solve the challenges faced in constructing a reliable Gaussian EP approximation for the analytically intractable likelihood resulting from the NN observation model by adopting the probabilistic backpropagation method by Jose and Ryan \cite{jose2015propagation}. \par Finally, we derive the VEP for the hierarchical Bayesian model with the Logistic regression as the likelihood and the Gaussian prior distribution. We work in the context of deep neural networks. Working with Logistic regression by adopting the approximate lower bound of the logistic function (Jaakkola and Jordan, $2000$) extends the work by Jose and Ryan, \cite{jose2015propagation}. However, this work computes the parameters of the hyperprior distribution using the marginal likelihood.
\section{Basic Description of EP Algorithms}
In this section, we briefly review the EP, and VI algorithms upon which our model is hinged. Let us observe a dataset consisting of the $N$ i.i.d. samples $\mathcal{D} = \{\vec{x}_i\}_{i-1}^N$ from the parametric probabilistic model $p(\vec{x}|\vec{\theta})$ parameterized by an unknown D-dimensional vector $\vec{\theta}$ whose distribution is $p_0(\vec{\theta})$. Exact Bayesian involves the computation of typically intractable posterior distribution over the parameters. Given the data the posterior distribution is formulated as follows;
\begin{equation}
	p(\vec{\theta}|\mathcal{D}) \propto \mathlarger \prod_{i=1}^N t(\vec{\theta}) p_0(\vec{\theta}) \approx q(\vec{\theta}) \propto p_0(\vec{\theta}) \mathlarger \prod_{i=1}^N \tilde{t}_i(\vec{\theta})
	\label{equ:1}
\end{equation}
where the $p(\vec{x}_i|\vec{\theta}) = t_i(\vec{\theta})$ is the exact term, $\tilde{t}(\vec{\theta})$ is the approximate terms, and $q(\vec{\theta})$ is the approximate posterior distribution that will be refined by EP. The goal of EP is to capture the contribution of data point through the likelihood term to the posterior by refining the approximate terms. i.e. $t_i(\vec{\theta}) \propto p(\vec{x}_i|\vec{\theta}).$ In the same spirit, one approach would be to search for the approximate posterior that minimizes the Kullback Leibler (KL) Divergence between the exact posterior and the distribution formed by the replacing one of the likelihood by the approximate term $t_i(\vec{\theta})$ i.e. $\text{KL}[p(\vec{\theta}|\mathcal{D})||p(\vec{\theta}|\mathcal{D})\tilde{t}_i(\vec{\theta})/p(\vec{x}_i|\vec{\theta})]$. However, the update still becomes intractable due to the involvement of the exact posterior distribution. EP rather approximates this procedure by removing one observation from the posterior i.e. $p_{-i}(\vec{\theta}) \propto p(\vec{\theta}|\vec{x}_i)/t_i(\vec{\theta})$ called the exact leave-one-out posterior. This is replaced by the approximate leave-one-out posterior $q_{-i}(\vec{\theta}) \propto q(\vec{\theta})/\tilde{t}(\vec{\theta})$ called cavity distribution. Since this couples the updates for the approximating factors, the updates must now be iterated. In a more general perspective, EP first selects the factor for update and eliminate it from the approximate posterior distribution to produce the cavity distribution. Then, there is an inclusion of the corresponding likelihood to produce the tilted posterior $\tilde{p}_i(\vec{\theta}) \propto q_{-i}(\vec{\theta})t_i(\vec{\theta})$. EP updates the approximate posterior distribution by minimizing the $\text{KL}[\tilde{p}(\vec{\theta})||q(\vec{\theta})]$. This establishes that the same measure of contribution of the exact term is made on both the exact posterior and tilted posterior. The KL minimization often reduces to moment matching when the approximate distribution is in the exponential family. Finally, we update the approximate term by dividing the old approximate posterior from the new approximate posterior. In practice, EP often performs well due to the local update. There is however one cogent limitation that complicates EP update. The computation of the normalizing factor by marginalizing the parameters. This often becomes intractable when there are many integration arising from too many parameters to marginalize. \par Having established EP with local update, another deterministic approximate method that has different viewpoint is the Variational Bayes (VB). Now, let us consider in more detail how the concept of variational optimization can be applied to the inference problem. Suppose we have a full Bayesian model in which all parameters are treated as random variables. Our probabilistic model specifies the joint distribution $p(\vec{x},\vec{\theta})$, and the goal is to find an approximation for the posterior distribution $p(\vec{\theta}||\vec{x})$ as well as for the model evidence $p(\vec{x})$. We minimize $\text{KL}[q(\vec{\theta})||p(\vec{\theta}||\vec{x})]$ (the reverse version of the KL for EP). Note that unlike the EP we previously explained, VI averages over the approximate posterior.
\section{Variational Expectation Propagation}
According to Minka \cite{minka2005ep}, EP minimizes the forward Kullback-Leibler divergence $\text{KL}(p||q)$ that uses the matching-moment if only the two distributions are in the same exponential family most importantly Gaussian family. However, many studies have brought into limelight how rigorous and intractable EP could be when the tilted and approximate distribution come from different family of distributions entirely. Zoeta and Heskes, $2005$ uses Gaussian Quadrature for the problem of mismatching moments in EP with different family of distributions such as Beta distributions. Also, Roberts and Olobatuyi, $2020$ (in press) uses the stochastic search algorithm to solve the problem of mismatching moments between Gaussian and Exponential distributions. On the contrary, VB minimizes the reverse Kullback-Leibler divergence $\text{KL}(q||p)$. VB has been used to handle many combinations of incongruous distributions by indirectly maximizing the lower bound $\mathcal{L}(q)$ through optimizing with respect to the distribution $q$. The difference between these two Kullback-Leibler divergences can be understood by noting that there is a large positive contribution to the Kullback-Leibler divergence $\text{KL}(q||p)$ from the region of the latent space in which the $p$ is near zero unless $q$ is also close to zero. Thus minimizing this form of KL divergence leads to distributions $q$ that avoid the region in which $p$ is small. Similar to this framework, Magnus et al., ($2009$) hybridized VB and EP for Bayesian sparse factor analysis. A comparison study carried out by Kim and Wand, ($2016$) on the accuracy power of the mean and variance estimates produced by both VB and EP algorithms shows that the mean estimate produced by VB tends to be more accurate than the mean estimate produced by EP algorithm. On the contrary, the variance estimate produced by VB is underestimated or less accurate than the variance estimate produced by EP algorithm. This also confirms the study by \cite{bishop2006} that shows how well approximated to the mean of the exact posterior distribution the mean of the approximate posterior produced by VB but underestimates the variance estimate. \par Here, we show that the Kullback-Leibler divergence $\text{KL}(p||q)$ is less than or equal to $\text{KL}(q||p)$ augmented by any constant and local optimization to reflect the Leave-One-Out (LOO) method in EP. i.e. 
\begin{equation}
	\text{KL}(p||q) \leq \text{KL}(q||p) + \text{constant}
	\label{equ:2}
\end{equation}
The accuracy of the pure mean-field solution, treating the latent variables as factorized variables by augmenting the reverse KL divergence. 
By minimizing the $\text{KL}(p||q)$ of Equation \ref{equ:2}, the normalization constant $Z$ of Equation \ref{equ:30} following the matching moments of Minka \cite{minka2001ep}, the Kullback-Leibler divergence between $p$ and $q^{new}$ can be then be obtained as a function of $m$, $v$, and the gradient of $\log Z$ with respect to these quantities, namely
\begin{equation}
	m^{new} = m_{-i} + v_{-i} \nabla_m \log Z
	\notag
\end{equation}
\begin{equation}
	v^{new} = v_{-i} - v_{-i}^2 \bigg[\bigg(\nabla_m \log Z\bigg)^2 - 2\nabla_v \log Z\bigg]
	\label{equ:3}
\end{equation}
We present the Bayesian linear regression model example in Bishop, $(2009)$. This example has been solved by the variational Bayes approach. The likelihood function for $\mathbf{w}$, and the prior over $\mathbf{w}$ are given by
\begin{equation}
	p(\mathbf{t}|\mathbf{w}) = \mathlarger \prod_{i=1}^{N} \mathcal{N}(t_i|\mathbf{w}^T\phi_i,\beta^{-1}) \hspace{0.05in} \text{and} \hspace{0.05in} p(\mathbf{w}|\alpha) = \mathcal{N}(\mathbf{w}|\mathbf{0},\alpha^{-1})
	\label{equ:4}
\end{equation}  
where $\vec{\phi}_i = \vec{\phi}(\mathbf{x}_i)$. The prior over $\alpha$ is given thus
\begin{equation}
	p(\alpha) = Gam(\alpha|a_0,b_0)
	\label{equ:5}
\end{equation}
First by using EP algorithm, we approach EP from $\text{KL}(\hat{p}_i\|q_i)$ and $\text{KL}(q_i||\hat{p}_i)$. Here as EP, we have the cavity distribution, approximate posterior, and the tilted posterior as follows
\begin{equation}
	q_{-i}(\mathbf{w}) = \mathcal{N}(\mathbf{w}|m_{-i},v_{-i}), \hspace{0.02in} q(\mathbf{w}) = \mathcal{N}(\mathbf{w}|\vec{m}_w,\vec{v}_w), \hspace{0.03in} \text{and} \hspace{0.03in}\hat{p}_i(\mathbf{w}) = \mathcal{N}(\mathbf{w}|\hat{m}_i,\hat{v}_i)
	\label{equ:6}
\end{equation}
we show that $\text{KL}(\hat{p}_i\|q_i) \equiv \text{KL}(q_i\|\hat{p}_i)$. However, it is clear from the symmetric property of Kullback-Leibler divergence that $\text{KL}(p\|q)$ $\not=$ $\text{KL}(q\|p)$. We proceed from the reverse $\text{KL}$ divergence first according to the conventional variational Bayes. 
\begin{equation}
	\ln q_i(\mathbf{w}) = E_\alpha\bigg[\ln q_{-i}(\mathbf{w})t_i(\mathbf{w})\bigg] = E_\alpha\bigg[\ln \mathcal{N}(\mathbf{w}|m_{-i},v_{-i})\mathcal{N}(t_i|\mathbf{w}^T\phi_i,\beta^{-1})\bigg] \label{equ:7}
\end{equation}
We note here that the expectation with respect to $q(\alpha)$ is constant and it becomes irrelevant which will be removed going forward. Now, the mean and the variance of $q(\mathbf{w})$ is as follows
\begin{equation}
	m_w = \bigg(m_{-i} + v_{-i}\vec{\phi}_it_i\beta\bigg)\bigg(1 + v_{-i}\vec{\phi}^T\vec{\phi}\beta\bigg)^{-1} \hspace{0.04in} \text{and} \hspace{0.04in} v_w = v_{-i}\bigg(1 + v_{-i}\vec{\phi}^T\vec{\phi}\beta\bigg)^{-1}
	\label{equ:8}
\end{equation} 
Now from the direction of forward Kullback-Leibler divergence, first we compute the normalizing constant
\begin{equation}
	Z_t = \mathlarger \int \mathcal{N}(\mathbf{w}|m_{-i},v_{-i})\mathcal{N}(t_i|\mathbf{w}^T\phi_i,\beta^{-1}) \hspace{0.02in} dw = \mathcal{N}(m_t, v_t)
	\label{equ:9}
\end{equation}
where the mean $m_t$ and variance $v_t$ of $t$ are as follows
\begin{equation}
	m_t = m_{-i} \vec{\phi}_i \hspace{0.05in} \text{and} \hspace{0.05in} v_t = \bigg(\beta^{-1} + v_{-1}\vec{\phi}^T\vec{\phi}\bigg)^{-1}
	\label{equ:10}
\end{equation} 
Computing $\nabla_m \log Z_t$ and $\nabla_v \log Z_t$ and using Equation \ref{equ:9} gives the following 
\begin{equation}
	m_w = \bigg(m_{-i} + v_{-i}\vec{\phi}_it_i\beta\bigg)\bigg(1 + v_{-i}\vec{\phi}^T\vec{\phi}\beta\bigg)^{-1} \hspace{0.04in} \text{and} \hspace{0.04in} v_w = v_{-i}\bigg(1 + v_{-i}\vec{\phi}^T\vec{\phi}\beta\bigg)^{-1}
	\label{equ:11}
\end{equation} 
This establishes the equivalence between $\text{KL}(\hat{p}_i\|q_i) \equiv \text{KL}(q_i\|\hat{p}_i)$ in a local approximation.
\section{The Model}
This section focuses on the multilayer perceptron NNs where the unknown function value $f_i = f(\mathbf{x}_i)$ related to a $d$- dimensional input vector $\mathbf{x}_i$ is modeled as 
\begin{equation}
	\hat{f}(\mathbf{x}_i) = \mathlarger \sum_{k=1}^{K}\mathbf{W}_{kL}^Tg(\mathbf{W}_{kl}^T\vec{z}_{l-1}), \hspace*{0.5in} l = 1,...,L
	\label{equ:12}
\end{equation} 
where $g(x)$ is a nonlinear activation function, $K$ the number of hidden units, and the $\mathcal{W} = \{\mathbf{W}_l\}_{l=1}^L$ is the collection or array of all the weights of the networks with dimension of $K_l \times (K_{l-1} + 1$) between the fully connected layers. We denote the output of the layers by by vectors $\{\vec{z}_l\}_{l=0}^L$ where $\vec{z}_0$ is the input layer. $\{\vec{z}_l\}_{l=1}^{L-1}$ represents the output of the hidden layer and $\vec{z}_L = \sigma(\hat{f}_i)$ is the output of the output layer. The activation functions for each hidden layer are Rectified Linear Units (RELUs) i.e., $a(x) = \max(x,0)$, (\cite{nair2010}). In the next subsection, we explain the likelihood function for the model.
\subsection{Likelihood Definitions}
Here, we illustrate the use of local variational methods for the Bayesian logistic regression model. This focuses on the variational treatment based on the approach of Jaakkola and Jordan $2000$. The variational treatment leads to the Gaussian approximation like the Laplace method. However, compared to the Laplace method, the greater flexibility of the variational approximation leads to improved accuracy. Furthermore, the variational approach can be as optimizing a well defined objective function given the rigorous bound on the model evidence. 
\subsection*{Binary-Class Classification}
Logistic regression has been treated from the standpoint of Monte Carlo sampling techniques (Dybowski and Roberts, $2005$). The output of the last layer is transformed using the sigmoid function for a binary output and softmax as a multiclass output. The variational approximation based on the lower bound allows the likelihood function for logistic regression, which is governed by the sigmoid or softmax to be approximated by the exponential of a quadratic form. \par We first note that the conditional distribution for $y$ can be written as 
\begin{equation}
	p(y_i|a_L,\vec{\Theta}) = \sigma(a_L)^{y_i} (1-\sigma(a_L))^{1-y_i}
	\notag
\end{equation}
\begin{equation}
	= e^{-y_ia_L} \frac{e^{-a_L}}{1 + e^{-a_L}} = e^{-y_ia_L} \sigma(-a_L)
	\label{equ:13}
\end{equation}
where $a_L = \hat{f}_i$ and the $\vec{\Theta}$ is the collection of all the hyperparameters and $\hat{f}$ is from the equation \ref{equ:12}. The variational lower bound on the logistic sigmoid function in \ref{equ:13} is given by 
\begin{equation}
	\sigma(u) \geq \sigma(\zeta) \exp \{(u-\zeta) / 2 - \lambda(\zeta) (u^2 - \zeta^2)\}
	\label{equ:14}
\end{equation}
where $\lambda(\zeta) = \frac{1}{2\zeta}\bigg[\sigma(\zeta) - \frac{1}{2}\bigg]$. Therefore, the likelihood function is written as 
\begin{equation}
	p(y_i|a_L, \vec{\Theta}) = e^{y_ia_L}\sigma(-a_L) \geq e^{y_ia_L} \sigma(\zeta) \exp \{-(a_L+\zeta) / 2 - \lambda(\zeta) ((a_L)^2 - \zeta^2)\}
	\label{equ:15}
\end{equation} 
Moreover, the bound is applied to each of the terms in the likelihood function separately, then there is a variational parameter $\zeta_i$ associated to each training set ($\mathbf{x}_i, y_i$). Finally, the lower bound for the likelihood function will be denoted as 
\begin{equation}
	h(\vec{\Theta}, \vec{\zeta}) = \mathlarger \prod_{i=1}^{N} \sigma(\zeta_i) \exp\bigg\{y_ia_L - (a_L + \zeta_i) / 2 - \lambda(\zeta_i)(a_L^2 - \zeta_i^2)\bigg\}
	\label{equ:16}
\end{equation}
The likelihood used is the lower bound of the Sigmoid function for binary classification which is presented in Equation \ref{equ:16} and this makes the posterior analytically intractable. 
\subsection{Prior Definitions}
We use the sparsity-promoting priors $p(w_{kjl}|\tau_{kjl})$ with hierarchical scale parameters $\tau^{-1}_{kjl}$ where the weight $w_{kjl}$ is the $k$:th row and $j$:th column of the $\mathbf{W}_l$, $\tau^{-1}_{kjl}$ controls the prior variance of all the weights $w_{kjl}$. We place a Gaussian prior over the weights as follows 
\begin{equation}
	p(w_{kjl}|\tau_{kjl}) = \mathcal{N}(w_{kjl}|0,\tau^{-1}_{kjl})
	\label{equ:17}
\end{equation}
where the variance is $\tau^{-1}_{kjl}$ in equation \ref{equ:17}. The grouping of the weights can be chosen freely and also other weight prior distribution can be used in place of Gaussian distribution. The approximate inference on the variance parameters $\vec{\tau}^{-1}_{l} > 0$ is carried out using non-negative supported prior distribution to constrain the variance to be non-negative. In doing so, the computationally most convenient alternative non-negative supported prior distribution is to employ rectified Gaussian prior distribution on the precision controlling parameter as follows;
\begin{equation}
	p(\tau_{kjl}) = \frac{2}{\text{erfc}(-m_0/\sqrt{v_0})}\mathcal{N}(\tau_{kjl} | m_0,v_0)U(\tau_{kjl})
	\label{equ:18}
\end{equation}
where $k = 1,...,K$, $j = 1,..., K_{l-1}+1$ and $l = 1,...,L$. Equation \ref{equ:18} corresponds to a rectified-Gaussian prior for the associated layer prior precision $\vec{\tau}_{l}$ and $U(.)$ is a step function, $m_0$ and $v_0$ are the location and scale parameter for the precision, respectively. It is easy to see that the rectified Gaussian prior is conjugate to a Gaussian likelihood and the posterior can be computed in the same manner as the standard Gaussian distribution since the $\text{erfc}(0) = 1$, then we have constant of $2$ in Equation \ref{equ:18}. \par However to solve the EP algorithm, the rectified Gaussian prior is only computationally possible if the location parameter $m_0$ is fixed to zero, making the \textit{erfc} function vanish. Also we note that the biases are already included in the setup of the weights matrices.
\subsection{The Posterior Distribution}
Given the previously explained prior definitions and a set of $N$ observations $\mathcal{D} = \{\mathbf{X}, \mathbf{y}\}$, where $\mathbf{y} = [y_i,...,y_N]^T$ and the features are $\mathbf{X} = [\mathbf{x}_1,...,\mathbf{x}_N]^T$, the joint posterior distribution of the prior and hyperparameters is as follows;
\begin{equation}
	p(\mathbf{w}, \vec{\tau}|\mathcal{D},\vec{\zeta},\vec{\gamma}) = Z^{-1}\mathlarger \prod_{i=1}^N h(y_i|\hat{f}_i,\zeta_i)\mathlarger \prod_{k=1}^{K_l}\mathlarger \prod_{j=1}^{K_{l-1}+1}\mathlarger \prod_{l=1}^{L} p(w_{kjl}|\tau_{kjl})\mathlarger \prod_{l=1}^{L} p(\vec{\tau}_l|\vec{\gamma})
	\label{equ:19}
\end{equation} 
where the $\vec{\gamma} = \{\zeta, m_0, v_0\}$ contains all the hyperparameters to be computed at the E-step of the EM version of EP, VB, and VEP algorithms, and $Z_{EP}$ is the approximation of the marginal likelihood $Z$ which is the marginal likelihood of the observations conditioned on $\vec{\gamma}$ as follows
\begin{equation}
	Z = p(\mathbf{y} | \mathbf{\mathbf{X}},\vec{\zeta},\vec{\gamma}) = \mathlarger \int h(\mathbf{y}|\mathbf{X},\mathbf{w},\vec{\zeta},\vec{\gamma}) p(\mathcal{W}|\vec{\tau}) p(\vec{\tau}|\vec{\gamma}) d\mathbf{w}d\vec{\tau}
	\label{equ:20}
\end{equation} 
\section{Approximate Inference}
In this section, we describe how approximate Bayesian inference on the unknown model parameters $\mathbf{w},\vec{\tau}$, and $\vec{\zeta}$ can be done efficiently using the variants of EP. First, in section \ref{sec:3.1}, we describe how the posterior approximation is formed using the approximate term and in section \ref{sec:3.2}, we discuss the hybridization of the VB and EP algorithm suitable for determining their parameters.
\subsection{The Approximate Posterior}\label{sec:3.1}
We form the analytically tractable approximation for the exact posterior distribution. We approximate all the likelihood and prior terms with unnormalized Gaussian distribution where appropriate. The Gaussian distribution has become a common used of approximating family for the weights of neural network, due to its matching moments nature.(Seeger, $2008$; Freitas, $1999$). However, we use the rectified Gaussian distribution for the prior distribution over the precision of the weights of the neural networks. This is important, as to place a nonzero constraint on the prior distribution. On the contrary, one could consider other exponential family distribution such as the gamma distribution for the weight precision parameter (Jose and Ryan, \cite{jose2015propagation}). We approximate the exact posterior distribution in Equation \ref{equ:900} as follows
\begin{equation}
	p(\mathbf{w}, \vec{\tau}|\mathcal{D},\vec{\zeta},\vec{\gamma}) = Z^{-1}\mathlarger \prod_{i=1}^N h(y_i|\hat{f}_i,\zeta_i)\mathlarger \prod_{k=1}^{K_l}\mathlarger \prod_{j=1}^{K_{l-1}+1}\mathlarger \prod_{l=1}^{L} p(w_{kjl}|\tau_{kjl}) p(\tau_{kjl}|\gamma_{jkl})
	\label{equ:21}
\end{equation} 
\begin{equation}
	\approx Z_{EP}^{-1}\mathlarger \prod_{i=1}^N \tilde{Z}_{y,i} \tilde{t}(\hat{f}_i)\mathlarger \prod_{k=1}^{K_l}\mathlarger \prod_{j=1}^{K_{l-1}+1}\mathlarger \prod_{l=1}^{L} \tilde{Z}_{kjl}^w\tilde{t}(w_{kjl})\tilde{t}(\tau_{kjl})
	\label{equ:22}
\end{equation} 
\subsection*{The Likelihood Term Approximations}
The exact likelihood terms that depend on the weights $\mathbf{w}$ through $\tilde{f}_i$ according to the Equation \ref{equ:22}
\begin{equation}
	h(y_i|a_L,\zeta_i)\approx \tilde{Z}_{y,i} \tilde{t}_i(a_L|\tilde{m}_i,\tilde{v}_i) = \tilde{Z}_{y,i}\mathcal{N}(a_L|\tilde{m}_i, \tilde{v}_i) 
	\label{equ:23}
\end{equation}
where $\tilde{Z}_{y,n} = \int \tilde{t}(a_L|\tilde{m}_i,\tilde{v}_i) \hspace{0.05in}da_L$ is a scalar scaling parameter or normalizing constant. Here, we have assumed that the all the weights are incorporated in $\mathbf{w}$ for both the hidden layer and the output layer. Note that the notation $\mathcal{N}$ is used for a normalized Gaussian distribution. Notice that we are approximating the lower bound to the sigmoid as the likelihood function that is the probability distribution which normalizes over the binary targets $y_i$, by an un-normalized Gaussian distribution over the latent variables $\hat{f}_i$. This is important because we are interested in how the likelihood behaves as a function of the latent $\hat{f}_i$. On the contrary, this is somewhat different from the regression setting which uses Gaussian distribution as the likelihood function and as linear model for the output $y_i$ which makes it a Gaussian distribution. We compute the posterior to investigate how the likelihood function behaves as a function of $\hat{f}_i$.
\subsection*{The prior Term Approximation}
The prior terms of the all the weights $w_{ijl}$ for $i = 1,...,K$, $j = 1,...,K_{l-1} + 1$ and $l = 1,...,L$ are approximated conventional by Gaussian distribution
We have used particularly a factorized distribution, due to the structure of the prior distribution over the precision of the weights, 
\begin{equation}
	p(w_{kjl}|\tau_{kjl}) \approx \tilde{Z}_{kjl}^w\tilde{t}(w_{kjl})\tilde{t}^{\tau}_{kjl}(\tau_{kjl}) \propto \mathcal{N}(w_{kjl}|\tilde{m}^w_{kjl},\tilde{v}^w_{kjl})\mathcal{N}(\tau_{kjl}|\tilde{m}^\tau_{kjl},\tilde{v}^\tau_{kjl})
	\label{equ:24}
\end{equation}
where a factorized site approximation with location and scale parameters $\tilde{m}_{kjl}^{w}$ and $\tilde{m}^{\tau}_{kjl}$, $\tilde{v}^{w}_{kjl}$, and $\tilde{v}^{\tau}_{kjl}$ are associated with the network weights and precision respectively. The approximation term for the $\tau_{kjl}$ is also assumed to be the rectified Gaussian distribution and any other exponential distribution could also be appropriate. 
\subsection*{The Joint Posterior Approximate}
The product of the independence local likelihood $\tilde{t}_i$ is 
\begin{equation}
	q(\vec{\hat{f}}) = \mathlarger \prod_{i=1}^N \tilde{Z}_{y,i} \mathcal{N}(\hat{f}_i|m_{\hat{f}},v_{\hat{f}}) = \mathcal{N}(\hat{\vec{f}}|\vec{m}_{\hat{f}}, \vec{v}_{\hat{f}}) \mathlarger \prod_{i=1}^{N} \tilde{Z}_{y,i}
	\label{equ:25}
\end{equation}
The prior and hyperprior that need to be processed multiple times using the expectation propagation are the factors in equation \ref{equ:22} as follows 
\begin{equation}
	q(\mathbf{w}, \vec{\tau}) = \mathlarger \prod_{k=1}^{K_l}\mathlarger \prod_{j=1}^{K_{l-1}+1}\mathlarger \prod_{l=1}^{L} \tilde{Z}_{kjl}^w\mathcal{N}(w_{kjl}|m_{kjl}^w,v_{kjl}^w)\mathcal{N}(\tau_{kjl}|m_{kjl}^{\tau},v_{kjl}^{\tau})
	\label{equ:26}
\end{equation}
In Equation \ref{equ:26}, we use the assumption of independence between the approximate posterior distributions and use the method of factorized distribution as follows
\begin{equation}
	q(\mathbf{w}, \vec{\tau}) = q(\mathbf{w})q(\vec{\tau})
	\label{equ:27}
\end{equation}
where the approximate posterior distribution for the network weights is given as follows 
\begin{equation}
	q(\mathbf{w}) = \mathlarger \prod_{k=1}^{K_l}\mathlarger \prod_{j=1}^{K_{l-1}+1}\mathlarger \prod_{l=1}^{L} \mathcal{N}(w_{kjl}|m_{kjl}^w,v_{kjl}^w)
	\label{equ:28}
\end{equation}
Conceptually, one can think of the approximate posterior distribution for the hyperprior $\tau$ in two ways, either by combining the approximate terms and the hyperprior distribution which gives the following 
\begin{equation}
	q(\vec{\tau}) = \mathlarger \prod_{k=1}^{K_l}\mathlarger \prod_{j=1}^{K_{l-1}+1}\mathlarger \prod_{l=1}^{L} \mathcal{N}(\tau_{kjl}|m_{kjl}^{\tau},{v}_{kjl}^{\tau}) 
	\label{equ:29}
\end{equation}
Now, multiplying the parameters approximation of $\mathbf{w}$ and $\vec{\tau}$ together with the prior in Equation \ref{equ:28} and \ref{equ:29} give the approximate posterior
\begin{equation}
	q(\mathbf{w}) \propto \mathcal{N}(\vec{M}_w, \vec{V}_{w}) \hspace*{0.2in} \text{and} \hspace*{0.2in} q(\vec{\tau}) \propto \mathcal{N}(\vec{M}_\tau, \vec{V}_{\tau})U(\tau)
	\label{equ:30}
\end{equation} 
where using the Gaussian multiplication strategy gives 
\begin{equation}
	\vec{M}_\tau = \vec{V}_\tau\tilde{\vec{V}}^{-1}_\tau \tilde{\vec{M}}_\tau \hspace*{0.2in} \text{and} \hspace*{0.2in} \vec{V}_\tau = \bigg(\vec{V}^{-1}_0 + \tilde{\vec{V}}^{-1}_\tau\bigg)^{-1}
	\label{equ:31}
\end{equation}
where the marginal posterior for the precision $\tau_{kjl}$ is given by 
\begin{equation}
	q(w_{kjl}) \propto \mathcal{N}(m^w_{kjl}, v^{w}_{kjl}) \hspace{0.2in} \text{and } \hspace{0.2in} q(\tau_{kjl}) \propto \mathcal{N}(m^\tau_{kjl}, v^{\tau}_{kjl})U(\tau)
	\label{equ:32}
\end{equation} 
where $m_{kjl}^{w}$ and $v_{kjl}^{w}$ are the mean and variance parameters for the approximate distribution of the network weights $q(w_{kjl})$ while $m_{kjl}^{\tau}$ and $v_{kjl}^{\tau}$ are the mean and variance parameters for the approximate distribution of the precision $q(\tau_{kjl})$. The mean vector $\vec{M}_{\tau}$ of the approximate posterior is the vector of $m_{kjl}^{\tau}$ and the covariance of the approximate posterior $\vec{V}_{\tau}$ is diagonal with $v_{kjl}^{\tau}$ for the approximate posterior.
\section{Hybridization of VB and EP}\label{sec:3.2}
The parameters of the local site approximations that define the approximate posterior distribution are determined using the hybridization of VB and EP. In the following, we give general description of the EP update for the likelihood and the weight prior terms. Here, we consider a sequentially updated EP.
\subsection{EP Update For the Hyperprior Terms}
As noted above in Equation \ref{equ:24}, each of the exact prior factors is approximated by a corresponding approximation prior give by 
\begin{equation}
	t(w_{kjl},\tau_{kjl}) = \mathcal{N}(w_{kjl}|0,\tau^{-1}_{kjl}) \hspace{0.2in} \text{and} \hspace{0.2in} \tilde{t}(\tau_{kjl}) = \mathcal{N}(\tau_{kjl}|\tilde{m}_{kjl}^{\tau},\tilde{v}_{kjl}^{\tau})
	\label{equ:33}
\end{equation}
First, we initialize all the $\tilde{t}(\tau_{kjl})$ uniformly, that is, $\tilde{m}^{\tau}_{kjl} = 0$ and $\tilde{v}^{\tau}_{kjl} = \infty$. EP starts to incorporate all the exact prior factors $t(w_{kjl},\tau_{kjl})$ into $q$ in $K \times (K_{l-1} + 1) \times L$ times. Here, we are interested in the individual precision parameter for each weights. This is relevant because it shows how accurate the weight estimates are at the update for each unit of every layer. The only demerit of this approach is the memory inefficiency. However, we don't store each update in memory. The first time $t(w_{kjl},\tau_{kjl})$ is incorporated into $q$, we update $\tilde{t}(\tau_{kjl})$ and $q$ as follows:
\begin{equation}
	\tilde{m}^\tau_{kjl} = 0 \hspace*{0.2in} \text{and} \hspace*{0.2in} \tilde{v}^\tau_{kjl} = v_0, \hspace*{0.2in}
	m^\tau_{kjl} = 0 \hspace*{0.2in} \text{and} \hspace*{0.2in} v^\tau_{kjl} = v_0
	\label{equ:34}
\end{equation}
where $v_0$ is the parameter of the rectified Gaussian hyperprior on $\tau$. On subsequent iterations, we refine the $\tilde{t}(\tau_{kjl})$ by first removing the approximate factor from the approximate posterior of $\tau$ to obtain the cavity distribution. This cavity distribution is computed as the fraction of the $q$ and $\tilde{t}$. The cavity marginal distribution on $\tau_{kjl}$ is therefore
\begin{equation}
	q_{-kjl}(\tau_{kjl}) = q(\tau_{kjl})\tilde{t}(\tau_{kjl})^{-1} = \mathcal{N}(\tau_{kjl}|m^\tau_{-kjl},v^\tau_{-kjl})
	\label{equ:35}
\end{equation}
where $m^\tau_{-kjl}$ and $v^\tau_{-kjl}$ are as follows:
\begin{equation}
	(v^\tau_{-kjl})^{-1} = (v^\tau_{kjl})^{-1} - (\tilde{v}^\tau_{kjl})^{-1}
	\notag
\end{equation}
\begin{equation}
	m^\tau_{-kjl} = m^\tau_{kjl} + (\tilde{v}^{\tau}_{kjl})^{-1} v^\tau_{-kjl} (m^\tau_{kjl} - \tilde{m}^\tau_{kjl})
	\label{equ:36}
\end{equation}
The cavity for the marginal distribution of $w_{kjl}$ is also 
\begin{equation}
	q_{-kjl}(w_{kjl}) = q(w_{kjl})\tilde{t}(w_{kjl})^{-1} = \mathcal{N}(w_{kjl}|m^w_{-kjl},v^w_{-kjl})
	\label{equ:35}
\end{equation}
where $m^w_{-kjl}$ and $v^w_{-kjl}$ are as follows:
\begin{equation}
	(v^w_{-kjl})^{-1} = (v^w_{kjl})^{-1} - (\tilde{v}^w_{kjl})^{-1}
	\notag
\end{equation}
\begin{equation}
	m^w_{-kjl} = m^w_{kjl} + (\tilde{v}^w_{kjl})^{-1} v^w_{-kjl} (m^w_{kjl} - \tilde{m}^w_{kjl})
	\label{equ:36}
\end{equation}
\subsection{Computing the Tilted for $\tau_{kjl}$, and $w_{kjl}$}
After incorporating all the prior factors, we compute the tilted posterior distribution $\hat{p}(\tau_{kjl})$. The tilted distribution is formed by combining the cavity with the exact prior term $t(\tau_{kjl})$:
\begin{equation}
	\hat{p}(\tau_{kjl}) = \hat{Z}_w^{-1} q_{-kjl}t(\tau_{kjl})p(\tau_{kjl}) = \mathcal{N}(\tau_{kjl}|\hat{m}^\tau_{kjl}, \hat{v}^\tau_{kjl})
	\label{equ:37}
\end{equation} 
where the normalizing factor $Z_w$ is given as follows
\begin{equation}
	Z_w = \mathlarger \int t(\tau_{kjl})p(\tau_{kjl})q_{-kjl}\hspace{0.02in}d\hspace{0.02in}\tau_{kjl} = \mathlarger \int \mathcal{N}(\tau_{kjl}|0,v^\tau_0)\mathcal{N}(\tau_{kjl}|m^\tau_{-kjl},v^\tau_{-kjl})U(\tau_{kjl})\hspace{0.02in}d\hspace{0.02in}\tau_{kjl}
	\label{equ:38}
\end{equation}
We compute the $\log Z_w$ from the $\text{KL}(q||p)$ of VB instead of a direct computation of $Z$ by $\text{KL}(p||q)$. We compute $\log Z_w$ from $\text{KL}(q||p)$, with $\theta_{kjl} = (w_{kjl},\tau_{kjl})$ as follows:
\begin{equation}
	-\text{KL}(q_{kjl}||\hat{p}_{kjl}) = \mathlarger \int q_{kjl}(\theta_{kjl}) \log \frac{\hat{p}_{kjl}(\theta_{kjl})}{q_{kjl}(\theta_{kjl})} \hspace{0.02in} d\theta_{kjl}
	\label{equ:39}
\end{equation}
\begin{equation}
	= \mathlarger \int q_{kjl}(\theta_{kjl}) \bigg[\log\bigg( \frac{{t}_{kjl}(\theta_{kjl})p(\tau_{kjl})q_{-kjl}(\theta_{kjl})}{Z_w}\bigg) - \log q_{kjl}(\theta_{kjl})\bigg] \hspace{0.02in} d\theta_{kjl}
	\label{equ:40}
\end{equation}
By rearranging Equation \ref{equ:40} we obtain
\begin{equation}
	-\text{KL}(q_{kjl}||\hat{p}_{kjl}) = \mathlarger \int q_{kjl}(\theta_{kjl}) \log\bigg( \frac{{t}_{kjl}(\theta_{kjl})p(\tau_{kjl})q_{-kjl}(\theta_{kjl})}{q_{kjl}(\theta_{kjl})}\bigg)\hspace{0.02in} d\theta_{kjl} - \log Z_w 
	\label{equ:41}
\end{equation}
Making the $\log Z_w$ the subject of the formula and rearranging we obtain
\begin{equation}
	\log Z_w = \mathlarger \int q_{kjl}(\theta_{kjl}) \log\bigg( \frac{{t}_{kjl}(\theta_{kjl})p(\tau_{kjl})q_{-kjl}(\theta_{kjl})}{q_{kjl}(\theta_{kjl})}\bigg)\hspace{0.02in} d\theta_{kjl} + \text{KL}(q_{kjl}||\hat{p}_{kjl})
	\label{equ:42}
\end{equation}
where 
\begin{equation}
	\mathcal{L}(\tau_{kjl}) = \mathlarger \int q_{kjl}(\theta_{kjl}) \log\bigg( \frac{{t}_{kjl}(\theta_{kjl})p(\tau_{kjl})q_{-kjl}(\theta_{kjl})}{q_{kjl}(\theta_{kjl})}\bigg)\hspace{0.02in} d\theta_{kjl}
	\label{equ:43}
\end{equation}
then using factorized method $q_{kjl}(\theta_{kjl}) = q_{kjl}(w_{kjl}, \tau_{kjl}) = q_{kjl}(w_{kjl})q_{kjl}(\tau_{kjl})$ 
\begin{equation}
	\log Z_w = \mathcal{L}(\theta_{kjl}) + \text{KL}(q_{kjl}||\hat{p}_{kjl})
	\label{equ:44}
\end{equation}
\subsection{The hyperprior parameters $\tau_{kjl}$}
Here, just like VB, we maximize the lower bound in equation \ref{equ:44} and we take the expectation with respect to the $q(w_{jkl})$ using the following factorized method.
Thus, minimizing Kullback-Leibler divergence is equivalent to maximizing the lower bound, we select all the exact distributions that depend on only $\tau_{kjl}$ and obtain a general expression for the optimal solution $q(\tau_{kjl})$ as follows
\begin{equation}
	\ln q^*(\tau_{kjl}) = \mathrm{E}_w\bigg[t(w_{kj},\tau_{kjl})p(\tau_{kjl})q_{-kjl}(\tau)\bigg] + \text{const}
	\label{equ:45}
\end{equation}
\begin{equation}
	m^\tau_{kjl} = v^\tau_{kjl} (v^\tau_{-kjl})^{-1} m^\tau_{-kjl} - \frac{1}{2} v^{\tau}_{kjl}(v^w_{kjl} + [m^w_{kjl}]^2)\hspace*{0.2in}
	\label{equ:46}
\end{equation}
where we have used the expectation with respect to $q_w$ and $\mathrm{E}(w^2) = v_w + m^2_w$ and Equation \ref{equ:46} becomes
\begin{equation}
	m^\tau_{kjl} = \bigg[m^\tau_{-kjl} - \frac{1}{2} v_{-kjl}^{\tau}(v^w_{kjl} + [m^w_{kjl}]^2)\bigg]\frac{v_0}{v_{-kjl}^{\tau} + v_0}\hspace*{0.2in} \text{and} \hspace{0.2in}v^\tau_{kjl} = \bigg(v^{-1}_0 + (v^\tau_{-kjl})^{-1}\bigg)^{-1}
	\label{equ:47}
\end{equation}
\subsection{The Prior weights $w_{kjl}$}
We compute the approximate posterior mean $m_w$ and variance $v_w$ according to the setup of variational expectation propagation. We use the expectation with respect to the approximate posterior $q(\tau_{kjl})$ and factorize all that depend only on the weights $w_{kjl}$ as follows
\begin{equation}
	\ln q^*(w_{kjl}) = \mathrm{E}_{\tau}\bigg[t(w_{kjl},\tau_{kjl})q_{-kjl}(w_{kjl})\bigg] + \text{const}
	\label{equ:48}
\end{equation}
then the mean and variance of approximate posterior $q(w_{kjl})$ are computed as 
\begin{equation}
	m^w_{kjl} = m^w_{-kjl} (v^w_{-kjl})^{-1} v^w_{kjl}\hspace*{0.2in} \text{and} \hspace{0.2in} v^w_{kjl} = \bigg[(v^w_{-kjl})^{-1} + m^\tau_{kjl}\bigg]^{-1}
	\label{equ:49}
\end{equation}
In Equation \ref{equ:49}, $\mathrm{E}[\tau_{kjl}] = m^{\tau}_{kjl}$.
Finally, we update the parameters of the approximate factor $\tilde{t}(\tau_{kjl})$ and $\tilde{t}(w_{kjl})$
\begin{equation}
	\tilde{v}^\tau_{kjl} = \bigg[(v_{kjl}^\tau)^{-1} - (v_{-kjl}^\tau)^{-1}\bigg]^{-1} \hspace*{0.2in} \text{and} \hspace{0.2in} \tilde{m}^\tau_{kjl} = \tilde{v}^\tau_{kjl}\bigg[m_{kjl}^{\tau}(v^{\tau}_{kjl})^{-1} - m_{-kjl}^{\tau}(v^\tau_{-kjl})^{-1}\bigg]
\end{equation}
\begin{equation}
	\tilde{v}^w_{kjl} = \bigg[(v_{kjl}^w)^{-1} - (v_{-kjl}^w)^{-1}\bigg]^{-1} \hspace*{0.2in} \text{and} \hspace{0.2in} \tilde{m}^w_{kjl} = \tilde{v}^w_{kjl}\bigg[m_{kjl}^{w}(v^{w}_{kjl})^{-1} - m_{-kjl}^{w}(v^w_{-kjl})^{-1}\bigg]
\end{equation}
respectively.
\subsection{EP Update For the Likelihood Terms}
Here, we consider the procedures for updating the likelihood sites $\tilde{t}(w_{kjl})$ and approximate posterior $q(w_{kjl})$ defined in Equation \ref{equ:23}. The exact likelihood terms $p(y_i|f_i)$ is a Logistic regression model and approximated by the lower bound from a Taylor series $h(\vec{\Theta},\vec{\zeta})$ which does not depend on the weight precision $\tau_{kjl}$. The posterior approximations can be factorized as $q(z_{kjl},a_{kjl},w_{kjl})$ and the exact likelihood is $p(y_i|a_L) = t(a_L)$ where $a_L$ is the matrix multiplication for the last layer.
Now, we compute the $Z_y$ as follows
\begin{equation}
	Z_y = \mathlarger \int h(y_i|a_L)\mathcal{N}(a_L|m^{a_L}_{kj},v_{kj}^{a_L}) \hspace{0.04in} d \hspace{0.01in}a_L
	\label{equ:50}
\end{equation}
\begin{equation}
	= \exp\biggl\{y_ia_L-(a_L-\zeta_i)/2 - \lambda(\zeta_i)(a_L^2-\zeta_i^2) - \frac{1}{2v_{kj}^{a_L}}(a_L^2 - m_{kj}^{a_L})^2\biggr\}
	\label{equ:51}
\end{equation}
by integrating out the matrix multiplication $a_L$, we have
\begin{equation}
	Z_y = \mathcal{N}(y_i|m_y, v_y)
	\label{equ:52}
\end{equation} 
where the mean $m_y$ and $v_y$ are as follows
\begin{equation}
	m_y = \frac{1}{2} - \frac{m^{a_L}_{kj}}{v^{a_L}_{kj}} \hspace{0.09in} \text{and} \hspace{0.09in} v_y = \frac{1}{\zeta_i}\bigg[\frac{1}{2} - \sigma(\zeta_i)\bigg] - \frac{1}{v_{kj}^{a_L}}
	\label{equ:53}
\end{equation} 
Note that we have made use of the $$\lambda(\zeta_i) = \frac{1}{2\zeta_i}\bigg[\sigma(\zeta_i) - \frac{1}{2}\bigg]$$
The updated rule for the mean and variance of the approximate posterior of $q(f_i) = q(a_L)$ in Equation \ref{equ:25} is given below
\begin{equation}
	m_{new}^{a_L} = m_{\text{old}}^{a_L} + [m_y - y_i]v_y^{-1} \hspace{0.04in} \text{and} \hspace{0.04in} v_{new}^{a_L} = v_{old}^{a_L} + v_y^{-1}[2m_{old}^{a_L}(y_i-m_y)-1]
	\label{equ:54}
\end{equation}

\section{Probabilistic Back-propagation}
In this section we describe a probabilistic back-propagation algorithm for this model. PBP does not use point estimates for the synaptic weights in the network Jose and Ryan \cite{jose2015propagation}. Instead, it uses a collection of one-dimensional Gaussian, each one approximating the marginal posterior distribution of a different weight. PBP also has two phases equivalent to the ones of BP. In the first phase, the input data is propagated forward through the network. However, since the weights are now random, the activation produced in each layer are also random and result in (intractable) distributions. PBP sequentially approximates each of these distributions with a collection of one-dimensional Gaussian that match their marginal means and variances. At the end of this phase, PBP computes, instead of the prediction error, the logarithm of the marginal probability of the target variable. In the second phase, the gradients of this quantity with respect to the means and variances of the approximate Gaussian posterior are propagated back using reverse-mode differentiation as in classic back-propagation. These derivatives are finally used to update the means and variances of the posterior approximation.
\subsection{Derivation of the gradients}
We derive the gradient of the gradient of the logarithm of the marginal likelihood, that is the $\log Z_y$ given in Equation \ref{equ:50}, with respect to the means and variance of the network weights in the Gaussian approximate posterior $q$. In PBP, the corresponding algorithm has two variables such as the means and variance for each neuron. The activation function used at each layer is the RELU activation and this becomes random since the weights are now random. The output of each layer is denoted as $z_l$ and the matrix multiplication is denoted by $a_l$ for $l = 1,...,L$. The activation function used for the last layer is the sigmoid function $\sigma(.)$. We start by propagating forward through the network from the input layer to the last layer called output layer. Let us assume for the moment that we have $L = 3$ before the general concept of PBP. 
\subsection{The Forward Propagation}
Consider a class of neural networks defined the function form 
\begin{equation}
	z_L = \frac{1}{1 + \exp[-a_L]}
	\label{equ:61}
\end{equation}
where $a_L$ stands the matrix multiplication of the last layer and it is explicitly written as follows
\begin{equation}
	a_L = \vec{w}_L^Tg_L\bigg[\sum_{l = 1}^{L-1}\sum_{k = 1}^{K}g_l\bigg(\sum_{j=1}^{J}w_{kjl}\vec{z}_{l-1}\bigg)\bigg] 
	\label{equ:62}
\end{equation}
\begin{equation}
	\text{For Layer} \hspace{0.05in} L=1, \hspace{0.05in} \mathrm{E}\big[\mathbf{w}_1^Tz_0\big] \hspace{0.05in} \text{and} \hspace{0.05in} m_{kj}^w m^{z_0}  
	\label{equ:63}
\end{equation}
\subsection{The Backpropagation}
For the last layer,
\begin{equation}
	\frac{\partial Z_y}{\partial m_{kj}^{w_L}} = \frac{\partial \log Z_y}{\partial m_{kj}^{a_L}}\frac{\partial m^{a_L}_{kj}}{\partial m_{kj}^{w_L}} + \frac{\partial \log Z_y}{\partial v_{kj}^{a_L}}\frac{\partial v^{a_L}_{kj}}{\partial m_{kj}^{w_L}}, \hspace{0.08in}\text{and} \hspace{0.08in}
	\frac{\partial Z_y}{\partial v_{kj}^{w_L}} = \frac{\partial \log Z_y}{\partial m_{kj}^{a_L}}\frac{\partial m^{a_L}_{kj}}{\partial v_{kj}^{w_L}} + \frac{\partial \log Z_y}{\partial v_{kj}^{a_L}}\frac{\partial v^{a_L}_{kj}}{\partial v_{kj}^{w_L}}
	\label{equ:64}
\end{equation}
For the hidden layers
\begin{equation}
	\frac{\partial Z_y}{\partial m_{kj}^{w_l}} = \frac{\partial \log Z_y}{\partial m_{kj}^{a_l}}\frac{\partial m^{a_l}_{kj}}{\partial m_{kj}^{w_l}} + \frac{\partial \log Z_y}{\partial v_{kj}^{a_l}}\frac{\partial v^{a_l}_{kj}}{\partial m_{kj}^{w_l}}, \hspace{0.08in} \text{and} \hspace{0.08in} \frac{\partial Z_y}{\partial v_{kj}^{w_l}} = \frac{\partial \log Z_y}{\partial m_{kj}^{a_l}}\frac{\partial m^{a_l}_{kj}}{\partial v_{kj}^{w_l}} + \frac{\partial \log Z_y}{\partial v_{kj}^{a_l}}\frac{\partial v^{a_{l}}_{kj}}{\partial v_{kj}^{w_l}}
	\label{equ:65}
\end{equation}
where
\begin{equation}
	\frac{\partial \log Z_y}{\partial m_{kj}^{a_l}} = \frac{\partial \log Z_y}{\partial m_{kj}^{a_{l+1}}} \frac{\partial m_{kj}^{a_{l+1}}}{\partial m_{kj}^{a_l}} + \frac{\partial \log Z_y}{\partial v_{kj}^{a_{l+1}}} \frac{\partial v_{kj}^{a_{l+1}}}{\partial m_{kj}^{a_l}}, 
	\label{equ:66}
\end{equation}
\begin{equation}
	\frac{\partial \log Z_y}{\partial v_{kj}^{a_l}} = \frac{\partial \log Z_y}{\partial m_{kj}^{a_{l+1}}} \frac{\partial m_{kj}^{a_{l+1}}}{\partial v_{kj}^{a_l}} + \frac{\partial \log Z_y}{\partial v_{kj}^{a_{l+1}}} \frac{\partial v_{kj}^{a_{l+1}}}{\partial v_{kj}^{a_l}}
	\label{equ:67}
\end{equation}
\begin{equation}
	\frac{\partial m_{kj}^{a_{l+1}}}{\partial m_{kj}^{a_l}} = \frac{\partial m_{kj}^{a_{l+1}}}{\partial m_{kj}^{z_l}} \frac{\partial m_{kj}^{z_l}}{\partial m_{kj}^{a_l}} + \frac{\partial m_{kj}^{a_{l+1}}}{\partial v_{kj}^{z_l}} \frac{\partial v_{kj}^{z_l}}{\partial m_{kj}^{a_l}}, \hspace{0.05in} \text{and} \hspace{0.05in} \frac{\partial m_{kj}^{a_{l+1}}}{\partial v_{kj}^{a_l}} = \frac{\partial m_{kj}^{a_{l+1}}}{\partial m_{kj}^{z_l}} \frac{\partial m_{kj}^{z_l}}{\partial v_{kj}^{a_l}} + \frac{\partial m_{kj}^{a_{l+1}}}{\partial v_{kj}^{z_l}} \frac{\partial v_{kj}^{z_l}}{\partial v_{kj}^{a_l}}
	\label{equ:68}
\end{equation}
\begin{equation}
	\frac{\partial v_{kj}^{a_{l+1}}}{\partial m_{kj}^{a_l}} = \frac{\partial v_{kj}^{a_{l+1}}}{\partial m_{kj}^{z_l}} \frac{\partial m_{kj}^{z_l}}{\partial m_{kj}^{a_l}} + \frac{\partial v_{kj}^{a_{l+1}}}{\partial v_{kj}^{z_l}} \frac{\partial v_{kj}^{z_l}}{\partial m_{kj}^{a_l}}, \hspace{0.05in} \text{and} \hspace{0.05in} \frac{\partial v_{kj}^{a_{l+1}}}{\partial v_{kj}^{a_l}} = \frac{\partial v_{kj}^{a_{l+1}}}{\partial m_{kj}^{z_l}} \frac{\partial m_{kj}^{z_l}}{\partial v_{kj}^{a_l}} + \frac{\partial v_{kj}^{a_{l+!}}}{\partial v_{kj}^{z_l}} \frac{\partial v_{kj}^{z_l}}{\partial v_{kj}^{a_l}}
	\label{equ:69}
\end{equation}
The mean and variance of the output of the matrix multiplication at each level are defined as $m_{kj}^{a_l}$ and $v_{kj}^{a_l}$ respectively. Also, the mean and variance of the activation function which becomes the input of the next layer are defined as $m_{kj}^{z_l}$ and $v_{kj}^{a_l}$ respectively. First, the matrix multiplication is randomized following from Equation \ref{equ:63} by computing the first and second moments as follows
\begin{equation}
	E[a_2] = E[w_{kj}^2]E[z^2] = (\text{Var}(w) + (E[w])^2)(\text{Var}(z) + (E[z])^2)
	\label{equ:70}
\end{equation}
The first and second moments are given below
\begin{equation}
	m_{kj}^{a_l} = m_{kj}^{z_{l-1}}m_{kj}^{w_l} 
	\label{equ:71}
\end{equation}
\begin{equation}
	v_{kj}^{a_l} = (m_{kj}^{z_{l-1}})^2v_{kj}^{w_l} + v_{kl}^{z_{l-1}}(m_{kj}^{w_l})^2 + v_{kj}^{z_{l-1}}v_{kj}^{w_l}
	\label{equ:72}
\end{equation}
The randomized RELU activation function is given as follows
\begin{equation}
	m_{kj}^{z_l} = \vec{\Phi}(\alpha_{kj}) \bigg[m_{kj}^{a_l} + \sqrt{v_{kj}^{a_l}}\gamma_{kj}\bigg]
	\label{equ:73}
\end{equation}
\begin{equation}
	v_{kj}^{z_l} = m_{kj}^{z_l} \bigg[m_{kj}^{a_l} + \sqrt{v_{kj}^{a_l}}\gamma_{kj}\bigg] \vec{\Phi}(-\alpha_{kj}) + \vec{\Phi}(\alpha_{kj})v_{kj}^{a_l}(1-\gamma_{kj}^2 - \gamma_{kj}\alpha_{kj})
	\label{equ:74}
\end{equation}
where $\gamma_{kj} = \phi(\alpha_{kj})/\vec{\Phi}(\alpha_{kj})$, $\alpha_{kj} = m_{kj}^{a_l} / \sqrt{v_{kj}^{a_l}}$ with $\phi$, and $\vec{\Phi}$ denote the standard Gaussian pdf and cdf respectively.
The gradients starting from the normalizing constant $Z_y$ in Equation \ref{equ:50} are as follows
\begin{equation}
	\frac{\partial\log Z_y}{\partial m^{a_L}_{kj}} = [m_y - y_i]v_y^{-1}(v_{kj}^{a_L})^{-1}
	\label{equ:75}
\end{equation}
\begin{equation}
	\frac{\partial\log Z_y}{\partial v^{a_L}_{kj}} = \frac{1}{(v^{a_L}_{kj})^2}\bigg[\frac{1}{2v_y^2}(y_i-m_y)^2 + \frac{m_{kj}^{a_L}}{v_y}(y_i-m_y) - \frac{1}{2v_y}\bigg]
	\label{equ:76}
\end{equation}
where Equation \ref{equ:50} is brought forward for convenience
\begin{equation}
	m_y = \frac{1}{2} - m^{a_L}_{kj} (v^{a_L}_{kj})^{-1} \hspace{0.09in} \text{and} \hspace{0.09in} v_y = \frac{1}{\zeta_i}\bigg[\frac{1}{2} - \sigma(\zeta_i)\bigg] - \frac{1}{v^{a_L}_{kj}}
	\label{equ:77}
\end{equation}
This is a gradient of the upper layer with respect to the lower layer of interest $a_{l}$ and $z_{l-1}$
\begin{equation}
	\frac{\partial m_{kj}^{a_l}}{\partial m_{kj}^{z_{l-1}}} = m_{kj}^{w_l}, \hspace{0.07in} \frac{\partial m_{kj}^{a_l}}{\partial v_{kj}^{z_{l-1}}} = 0, \hspace{0.07in} \frac{\partial v_{kj}^{a_l}}{\partial m_{kj}^{z_{l-1}}} = 2m_{kj}^{z_{l-1}}v_{kj}^{w_l}, \hspace{0.07in} \text{and} \hspace{0.07in} \frac{\partial v_{kj}^{a_l}}{\partial v_{kj}^{z_{l-1}}} = (m_{kj}^{w_l})^2 + v_{kj}^{w_l}.  
	\label{equ:78}
\end{equation}
This is a gradient with respect to the same layer of interest $a_l$ and $w_l$
\begin{equation}
	\frac{\partial m_{kj}^{a_l}}{\partial m_{kj}^{w_{l}}} = m_{kj}^{z_{l-1}}, \hspace{0.07in} \frac{\partial m_{kj}^{a_l}}{\partial v_{kj}^{w_{l}}} = 0, \hspace{0.07in} \frac{\partial v_{kj}^{a_l}}{\partial m_{kj}^{w_{l}}} = 2v_{kj}^{z_{l-1}}m_{kj}^{w_l}, \hspace{0.07in} \text{and} \hspace{0.07in} \frac{\partial v_{kj}^{a_l}}{\partial v_{kj}^{w_{l}}} = (m_{kj}^{z_{l-1}})^2 + v_{kj}^{z_{l-1}}.  
	\label{equ:79}
\end{equation}
This is a gradient with respect to the same layer of interest $z_l$ and $a_l$
\begin{equation}
	\frac{\partial m^{z_l}}{\partial m^{a_l}} = \vec{\Phi}(\alpha_{kj})\bigg[1+\sqrt{v_{kj}^{a_l}}\frac{\partial \gamma_{kj}}{\partial m_{kj}^{a_l}}\bigg] + \frac{\partial \alpha_{kj}}{\partial m_{kj}^{a_l}} \bigg[m_{kj}^{a_l}+\sqrt{v_{kj}^{a_l}}\gamma_{kj}\bigg] \phi(\alpha_{kj})
	\label{equ:80}
\end{equation}
\begin{equation}
	\frac{\partial m_{kj}^{z_l}}{\partial v_{kj}^{a_l}} = \frac{\partial \alpha_{kj}}{\partial v_{kj}^{a_l}} \bigg[m_{kj}^{a_l}+\sqrt{v_{kj}^{a_l}}\gamma_{kj}\bigg]\phi(\alpha_{kj}) + \vec{\Phi}(\alpha_{kj})\bigg[\sqrt{v_{kj}^{a_l}}\frac{\partial \gamma_{kj}}{\partial v_{kj}^{a_l}} + \frac{\gamma_{kj}}{2\sqrt{v_{kj}^{a_l}}}\bigg]
	\label{equ:81}
\end{equation}
\begin{equation}
	\frac{\partial v_{kj}^{z_l}}{\partial m_{kj}^{a_l}} = m_{kj}^{z_l}\bigg[1+\sqrt{v_{kj}^{a_l}}\frac{\partial \gamma_{kj}}{\partial m_{kj}^{a_l}}\bigg]\vec{\Phi}(-\alpha_{kj}) + \frac{\partial \alpha_{kj}}{\partial m_{kj}^{a_l}} \phi (\alpha_{kj}) v_{kj}^{a_l}(1-\gamma_{kj}^2 - \alpha_{kj}\gamma_{kj})
	\notag
\end{equation}
\begin{equation}
	-\bigg[m_{kj}^{a_l}+\sqrt{v_{kj}^{a_l}}\gamma_{kj}\bigg]\bigg[m_{kj}^{z_l}\phi(\alpha_{kj})\frac{\partial \alpha_{kj}}{\partial m_{kj}^{a_l}} - \vec{\Phi}(-\alpha_{kj}) \frac{\partial m_{kj}^{z_l}}{\partial m_{kj}^{a_l}}\bigg]
	\notag
\end{equation}
\begin{equation}
	- \vec{\Phi}(\alpha_{kj})v_{kj}^{a_l}\bigg[2\gamma_{kj}\frac{\partial \gamma_{kl}}{\partial m_{kj}^{a_l}} + \alpha_{kj}\frac{\partial \gamma_{kj}}{\partial m_{kj}^{a_l}} + \gamma_{kj}\frac{\partial \alpha_{kj}}{\partial m_{kj}^{a_l}}\bigg]
	\label{equ:82}
\end{equation}
\begin{equation}
	\frac{\partial v_{kj}^{z_l}}{\partial v_{kj}^{a_l}} = \vec{\Phi}(\alpha_{kj})\biggl\{\bigg[1-\gamma_{kj}^2-\alpha_{kj}\gamma_{kj}\bigg]\bigg[1+v_{kj}^{a_l}\gamma_{kj}\frac{\partial \alpha_{kj}}{\partial v_{kj}^{a_l}}\bigg] -v_{kj}^{a_l}\bigg[2\gamma_{kj}\frac{\partial \gamma_{kl}}{\partial v_{kj}^{a_l}} + \alpha_{kj}\frac{\partial \gamma_{kj}}{\partial v_{kj}^{a_l}} + \gamma_{kj}\frac{\partial \alpha_{kj}}{\partial v_{kj}^{a_l}}\bigg]\biggr\}
	\notag
\end{equation}
\begin{equation}
	+m_{kj}^{z_l}\biggl\{\bigg[\sqrt{v_{kj}^{a_l}}\frac{\partial \gamma_{kj}}{\partial v_{kj}^{a_l}} + \frac{\gamma_{kj}}{2\sqrt{v_{kj}^{a_l}}}\bigg]\vec{\Phi}(-\alpha_{kj}) - \bigg[m_{kj}^{a_l}+\sqrt{v_{kj}^{a_l}}\gamma_{kj}\bigg]\bigg[\phi(\alpha_{kj})\frac{\partial \alpha_{kj}}{\partial v_{kj}^{a_l}} - \frac{\vec{\Phi}(-\alpha_{kj})}{m_{kj}^{z_l}}\frac{\partial m_{kj}^{z_l}}{\partial v_{kj}^{a_l}}\bigg]\biggr\}
	\label{equ:83}
\end{equation}

we now compute the $\gamma$ and $\alpha$ with respect to $m_{kj}^{a_l}$ and $v_{kj}^{a_l}$
\begin{equation}
	\frac{\partial \gamma_{kj}}{\partial m_{kj}^{a_l}} = -\frac{\partial \alpha_{kj}}{\partial m_{kj}^{a_l}}\bigg[\alpha_{kj}\gamma_{kj}(\alpha_{kj}) + \gamma^2_{kj}(\alpha_{kj})\bigg] \hspace{0.2in} \text{and} \hspace{0.2in} \frac{\partial\alpha_{kj}}{\partial m_{kj}^{a_l}} = \frac{1}{\sqrt {v^{a_l}_{kj}}}
	\label{equ:84}
\end{equation}
\begin{equation}
	\frac{\partial \gamma_{kj}}{\partial v_{kj}^{a_l}} = -\frac{\partial \alpha_{kj}}{\partial v_{kj}^{a_l}}\bigg[\alpha_{kj}\gamma_{kj}(\alpha_{kj}) + \gamma^2_{kj}(\alpha_{kj})\bigg] \hspace{0.2in} \text{and} \hspace{0.2in} \frac{\partial\alpha_{kj}}{\partial v_{kj}^{a_l}} = -\frac{m_{kj}^{a_l}}{2v^{a_l}_{kj}\sqrt{v^{a_l}_{kj}}}
	\label{equ:85}
\end{equation}
\bibliographystyle{apalike}
\bibliography{vbep}

\end{document}